\documentclass{article}

    \PassOptionsToPackage{numbers, sort&compress}{natbib}

 \usepackage[preprint]{neurips_2026}

\usepackage[utf8]{inputenc} 
\usepackage[T1]{fontenc}    
\usepackage{hyperref}       
\usepackage{url}            
\usepackage{booktabs}       
\usepackage{amsfonts}       
\usepackage{nicefrac}       
\usepackage{microtype}      
\usepackage{xcolor}         
\usepackage{graphicx}
\usepackage{amsmath,amssymb}
\usepackage{wrapfig}
\usepackage{subcaption}

\newcommand{\rev}[1]{#1}

\title{TB-AVA: Text as a Semantic Bridge \\for Audio-Visual Parameter Efficient Finetuning}

\author{%
  \textbf{Seongah Kim} \quad \textbf{Dinh Phu Tran} \quad \textbf{Hyeontaek Hwang} \\
  \textbf{Saad Wazir} \quad \textbf{Duc Do Minh} \quad \textbf{Daeyoung Kim}\thanks{Corresponding author.} \\
  AI$^\text{2}$ Lab, KAIST \\
  \texttt{\{kimsa0322, kimd\}@kaist.ac.kr}
}

\begin{document}

\maketitle

\begin{abstract}
Audio-visual understanding requires effective alignment between 
heterogeneous modalities, yet cross-modal correspondence remains 
challenging when temporally aligned audio and visual signals lack clear 
semantic correspondence.
We propose to use text as a semantic anchor for audio-visual 
representation learning.
To this end, we introduce a parameter-efficient adaptation framework
built on frozen audio and visual encoders, centered on
\textbf{Text-Bridged Audio-Visual Adapter (TB-AVA)}, which enables
text-mediated interaction between audio and visual streams.
At the core of TB-AVA, \textbf{Gated Semantic Modulation (GSM)} selectively modulates feature channels based on text-inferred semantic
relevance. 
We evaluate the proposed approach on multiple benchmarks, including 
AVE, AVS, and AVVP, where the proposed framework achieves 
state-of-the-art performance, demonstrating text as an effective semantic anchor for 
parameter-efficient fine-tuning (PEFT) in audio-visual learning.
\end{abstract}

\section{Introduction}
\label{sec:intro}

Pretrained encoders for audio~\cite{chen2022htsat, BEATs_ICML23} and 
vision~\cite{liu2022swinv2} now provide strong unimodal features that 
transfer broadly across downstream tasks. Combining these representations to reason about the joint 
structure of audio and vision has proven considerably harder. In real-world 
video, sources can be off-screen and visual entities silent, so audible events 
and visible entities do not align frame by frame. In tasks such as event 
localization~\cite{tian2018ave}, video parsing~\cite{tian2020unified}, and 
segmentation~\cite{zhou2022avs}, the model must align audible events and visible entities to each other, 
inferring this from weak co-occurrence signals alone.

\begin{figure}[t]
  \centering
  \scalebox{0.85}{
  \includegraphics[width=\linewidth]{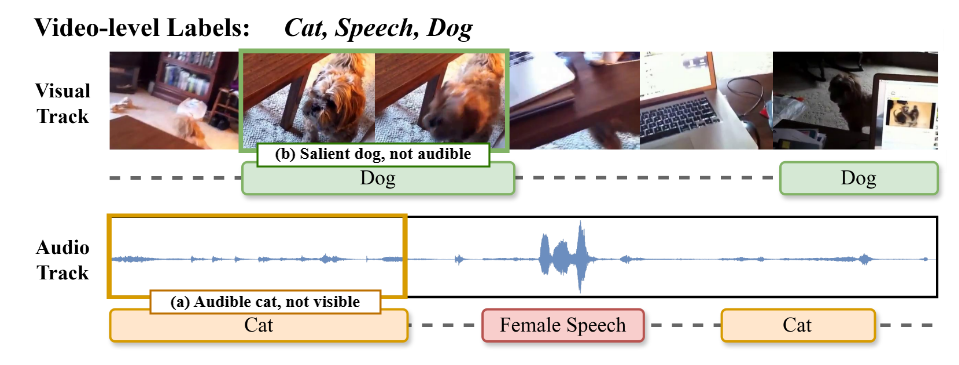}}
  \caption{Audio-visual ambiguity under weak supervision. 
The sounding cat is off-screen (a) and a non-sounding dog is 
visible (b). Temporal co-occurrence alone binds the wrong pair, 
motivating an external reference signal that is independent of 
modality-specific temporal alignment.}
  \label{fig:intro}
\end{figure}

The dominant alignment cue in current systems is \emph{temporal 
co-occurrence}~\cite{cheng2022jomold}: audio and visual features that 
fire simultaneously are bound together. This cue is reliable only when temporal overlap implies semantic 
correspondence, but this implication breaks under such video-level 
supervision~\cite{tian2020unified, lai2023valor}.
As shown in Figure~\ref{fig:intro}, the cat is audible but off-screen 
(a), while a dog is visible but silent (b). A baseline aligning audio and 
vision through temporal co-occurrence alone is structurally biased to 
bind the audible cat to the visible dog. The failure here cannot 
be fixed by larger models or more training data~\cite{mo2022closerlook}: 
it is a property of the inductive bias. Temporal co-occurrence fails 
whenever audible events and visible entities do not match one-to-one in 
each frame.

Resolving this ambiguity therefore requires a \emph{third signal} against which audio and visual evidence can be verified, yet existing approaches do not supply such a signal in a form that a general adapter over frozen encoders can use as a reference. Methods that bring text into audio-visual learning, such as T-VSL~\cite{TVSL_CVPR24} and TAViS~\cite{luo2025tavis}, tie text to a specific tri-modal backbone or foundation-model pair and to a single downstream task, rather than exposing it as a signal on top of arbitrary frozen encoders. Methods that operate without text on frozen encoders, such as AVMoE~\cite{AVMoE_NeurIPS24} and MoLT~\cite{rho2025moltmixturelayerwisetokens}, learn alignment implicitly through cross-modal fusion, with no reference signal to verify it. Methods that rely on a multimodal large model, such as 
Qwen2.5-Omni~\cite{Qwen25Omni_2025} and 
VideoLLaMA2~\cite{VideoLLaMA2_2024}, let cross-modal alignment emerge 
as a side effect of language modeling. However, they fail to maintain fine-grained 
audio-visual correspondence at the segment level~\cite{OVAVEL_CVPR25}.

What is missing is an \emph{explicit} semantic anchor that lives within the audio-visual representation, decoupled from backbone or downstream task, and exposing alignment as a controllable architectural component. A line of parameter-efficient adaptation in NLP and vision-language, from adapter tuning~\cite{NLPAdapter_ICML19} to context-conditioned prompting (CoOp)~\cite{CoOP_IJCV22}, demonstrates that text can play exactly this role when injected into model computation.

We bring this primitive into audio-visual learning, where, unlike in NLP and vision-language, the cross-modal correspondence to be aligned is itself unreliable. Our key idea is that text, encoded by a frozen text encoder over fixed 
class-level descriptions, can serve as a stable and verifiable semantic 
reference that disambiguates audio-visual correspondence at intermediate 
layers of frozen encoders, without modifying either backbone. We 
instantiate this idea in \emph{Text-Bridged Audio-Visual Adapter 
(TB-AVA)}, a parameter-efficient adapter that conditions cross-modal 
interaction on text embeddings. At the core of TB-AVA is 
\emph{Gated Semantic Modulation (GSM)}, a text-conditioned gated 
residual that selectively modulates only the dimensions where text is 
reliable, with channel-wise coefficients that are directly inspectable. It thereby aligns audio and visual features through a shared 
semantic basis, without re-introducing the ambiguity that text was 
meant to resolve.

Our contributions are:

\begin{itemize}
    \item We identify temporal co-occurrence as a primarily 
    limited inductive bias under weak supervision, and propose text 
    as a principled alignment primitive for frozen audio-visual 
    encoders, reframing text not as a task-specific auxiliary signal, 
    but as a backbone-agnostic semantic anchor that resolves 
    audio-visual ambiguity without modifying any encoder.
    
    \item We instantiate this perspective in TB-AVA, a 
    parameter-efficient adapter framework whose core module, GSM, 
    performs text-conditioned channel-wise gating that injects 
    text only where it is semantically reliable and suppresses 
    it elsewhere.
    
    \item Across three benchmarks, the proposed framework achieves state-of-the-art performance on AVE and strong competitive results on AVVP and AVS, demonstrating the effectiveness of text as a semantic anchor for parameter-efficient audio-visual alignment.
\end{itemize}

\section{Related work}
\label{sec:RW}

\begin{figure*}
  \centering
  \includegraphics[width=\textwidth]{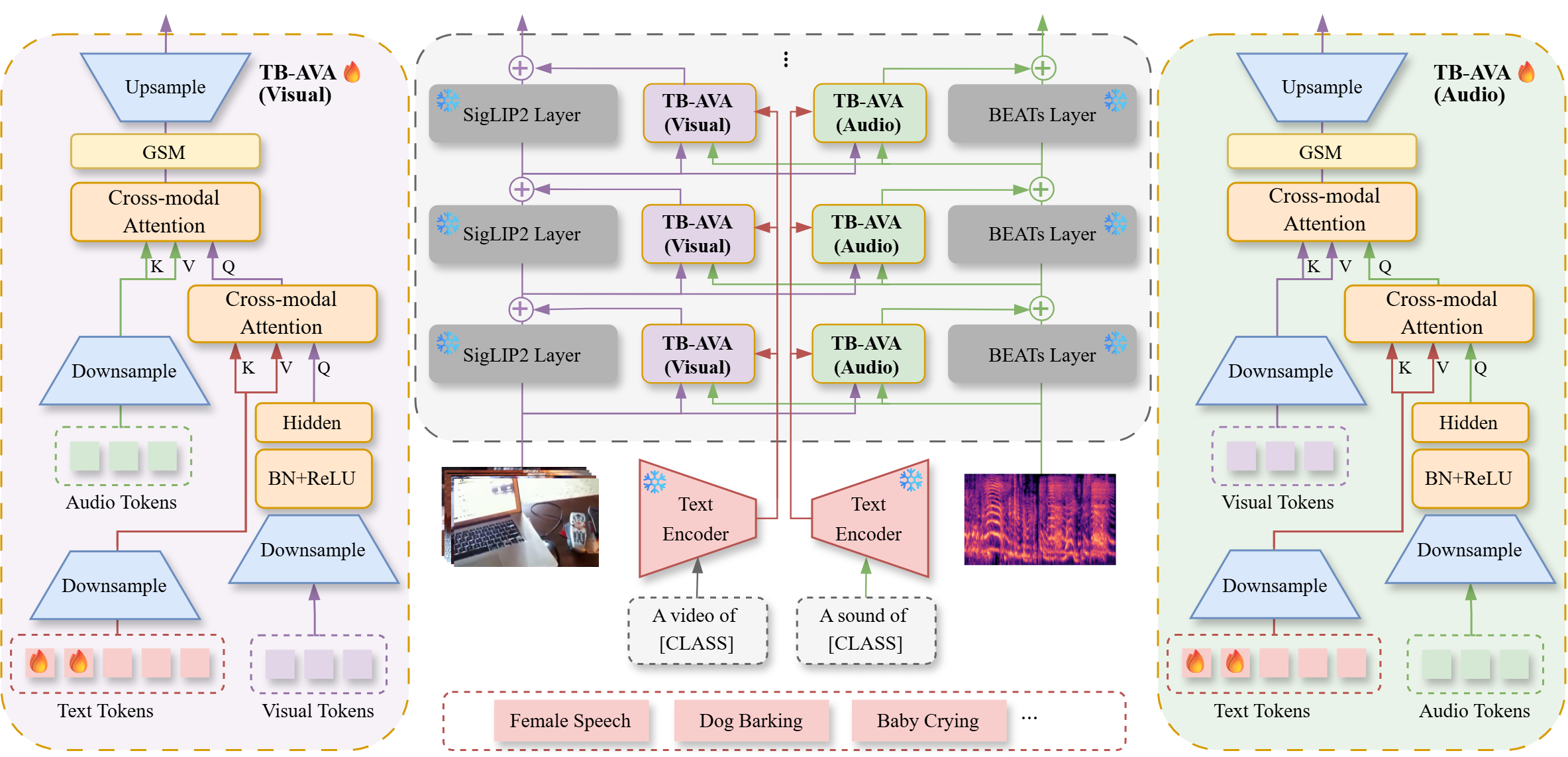}
  \caption{Architecture of TB-AVA. 
Lightweight adapters inserted between the first 12 layers of frozen 
SigLIP2 (visual, text) and BEATs (audio) encoders route cross-modal 
interaction through modality-matched text embeddings $T_v, T_a$ via 
Gated Semantic Modulation (GSM, Figure~\ref{fig:gsm}).}
  \label{fig:overview}
\end{figure*}

\noindent\textbf{Adapter-based audio-visual methods on frozen encoders.} 
Recent methods build on pretrained audio and visual transformers with 
lightweight cross-modal modules. LAVisH~\cite{LAVisH_CVPR23} inserts 
cross-modal adapter blocks between transformer layers, using a small 
set of latent tokens as an attention bottleneck for bidirectional 
fusion. DG-SCT~\cite{DGSCT_NeurIPS23} treats the counterpart modality 
as a soft prompt and decomposes interaction into spatial, channel, 
and temporal pathways with dynamic gating. AVMoE~\cite{AVMoE_NeurIPS24} 
introduces unimodal and cross-modal adapters as multiple experts and 
uses a lightweight router to allocate weights across them. More 
recent work pursues memory-efficient designs: 
Mettle~\cite{DBLP:journals/pami/ZhouLYZGLMHCW26} distills features 
from each frozen transformer layer into compact meta-tokens in 
parallel, bypassing backpropagation through the backbone, while 
MoLT~\cite{rho2025moltmixturelayerwisetokens} extracts layer-wise 
tokens via modality-specific and cross-modal adapter types and fuses 
them through a token fusion module. Across these designs, cross-modal 
interaction is a generic fusion or routing operator with no external 
reference signal, leaving alignment to emerge from weak supervision 
alone. Our adapter instead injects a text-derived semantic reference 
directly into the cross-modal interaction.

\noindent\textbf{Text-conditioned audio-visual methods.} 
Beyond fusion-based designs, several methods bring text into 
audio-visual learning. AudioCLIP~\cite{Audioclip_ICASSP22} extends 
contrastive language-image pretraining to a tri-modal 
audio-image-text embedding space, providing a foundation encoder 
rather than a downstream alignment mechanism. Built on this, 
T-VSL~\cite{TVSL_CVPR24} predicts the class of sounding entities 
and uses AudioCLIP text features as cross-attention conditioning 
for multi-source localization. TAViS~\cite{luo2025tavis} couples 
ImageBind for cross-modal alignment with SAM2 for segmentation, 
using pseudo text as class prototypes with audio-to-text and 
image-to-text alignment losses. Each is tied to a specific tri-modal 
backbone or foundation-model pair and to a single downstream task. 
We instead expose text as a backbone-agnostic, task-agnostic 
adaptation primitive applicable across AVE, AVVP, and AVS within 
a single framework.

\noindent\textbf{Text as an adaptation primitive in NLP and vision-language.} 
In NLP and vision-language, text-conditioned adaptation has become 
a well-established primitive for adapting frozen pretrained models 
without full fine-tuning. Adapter tuning~\cite{NLPAdapter_ICML19, AdapterFusion_EACL21} 
inserts trainable bottlenecks into each transformer layer; prefix 
and prompt tuning~\cite{li2021prefixtuning, lester2021prompttuning} 
prepend learned tokens to the input; conditional 
prompting~\cite{CoOP_IJCV22, zhou2022cocoop} treats class-level text 
descriptions as the conditioning signal. These approaches establish text as a stable conditioning signal in 
NLP and vision-language, but in audio-visual learning the modalities 
themselves are weakly aligned, leaving the role of text as an 
adaptation primitive.

We introduce TB-AVA, a parameter-efficient framework with 
\emph{Gated Semantic Modulation (GSM)} at its core, bridging audio 
and visual streams via a frozen text encoder over class-level 
descriptions. GSM selectively injects text-derived context only 
along channels where the reference is reliable, yielding explicit, 
controllable alignment over arbitrary frozen encoders.

\section{Method}

Our framework consists of frozen audio, visual, and text encoders, 
with lightweight TB-AVA inserted between backbone 
layers to enable text-bridged cross-modal interaction 
(Figure~\ref{fig:overview}). At the heart of each adapter is 
Gated Semantic Modulation (GSM), a text-conditioned 
channel-wise gating mechanism that selectively controls how cross-modal 
context is injected into each modality.

\subsection{Input representations}
\label{sec:inputs}

\paragraph{Visual.}
We represent each video as a sequence of $T$ RGB frames sampled at 
uniformly spaced timestamps, denoted as 
$\{\mathbf{V}_t\}_{t=1}^T$, where each frame 
$\mathbf{V}_t \in \mathbb{R}^{H \times W \times 3}$. Each frame is 
independently encoded by a frozen SigLIP2 visual 
encoder~\cite{SigLip_ICCV23}, chosen for its text-aligned visual 
representations. The visual representation at layer $l$ is denoted as 
$\mathbf{V}^l \in \mathbb{R}^{T \times N_v \times D_v}$, where $N_v$ is 
the number of patch tokens per frame and $D_v$ is the embedding 
dimension.

\paragraph{Audio.}
We segment the raw audio waveform into $T$ non-overlapping windows 
aligned with the visual frames, resample each to 16 kHz, and convert 
it into a log-mel spectrogram representation. The normalized 
spectrograms are fed into a frozen BEATs 
encoder~\cite{BEATs_ICML23}, which produces acoustic tokens. At layer $l$, the encoder outputs contextualized audio 
features $\mathbf{A}^l \in \mathbb{R}^{T \times N_a \times D_a}$, where 
$N_a$ denotes the number of acoustic tokens per window and $D_a$ is 
the embedding dimension.

\paragraph{Text.}
Each audio-visual sample is associated with textual class 
descriptions from the benchmark's event categories 
(e.g., ``female speech'' or ``dog barking''). To match the modality 
of each stream, we apply prompt templates: 
``a video of \{class\}'' for the visual stream and 
``a sound of \{class\}'' for the audio stream. The class-wrapped 
prompts are concatenated into a single text input per modality and 
encoded by the frozen SigLIP2 text encoder~\cite{SigLip_ICCV23}, 
which shares its embedding space with the visual encoder by 
construction. To allow the text representation to be adapted 
to the audio-visual learning context, we prepend $K$ learnable soft 
prompt tokens to each encoded sequence, with separate prompt sets for 
the visual and audio streams. This produces visual- and 
audio-conditioned text embeddings $\mathbf{T}_v, \mathbf{T}_a \in 
\mathbb{R}^{(K+L) \times D_t}$, where $L$ is the number of text 
tokens and $D_t$ is the embedding dimension. Only the soft prompts 
are trainable, while the text encoder remains frozen.

\subsection{Text-bridged audio-visual cross-modal adapter (TB-AVA)}
\label{sec:tbava}
We insert lightweight TB-AVA adapters between transformer 
layers of the frozen SigLIP2 visual encoder and the frozen BEATs audio 
encoder, allowing audio and visual representations to interact through 
text without modifying either backbone. Adapters are inserted into the 
first 12 layers of each encoder, where representations remain 
semantically flexible.

Each adapter operates symmetrically on both streams. We describe the 
visual stream below; the audio stream is symmetric. At layer $l$, the 
visual adapter receives the visual feature 
$\mathbf{V}^l \in \mathbb{R}^{T \times N_v \times D_v}$, the audio 
feature $\mathbf{A}^l \in \mathbb{R}^{T \times N_a \times D_a}$, and 
the modality-matched text embedding 
$\mathbf{T}_v \in \mathbb{R}^{L \times D_t}$. All three are projected 
into a low-dimensional bottleneck space, producing $\mathbf{z}_v$, 
$\mathbf{z}_a$, and $\mathbf{z}_t^v$.

The adapter retrieves cross-modal context in two steps, using text as 
the explicit anchor that mediates audio-visual interaction. First, the 
visual bottleneck attends over text to produce a text-aware visual 
representation:
\begin{equation}
\mathbf{c}_t^v = \text{Attention}(\mathbf{Q} = \mathbf{z}_v, 
\mathbf{K} = \mathbf{z}_t^v, \mathbf{V} = \mathbf{z}_t^v).
\end{equation}
This $\mathbf{c}_t^v$ then serves as the query that retrieves audio 
context aligned with the text-induced semantics:
\begin{equation}
\mathbf{c}_a = \text{Attention}(\mathbf{Q} = \mathbf{c}_t^v, 
\mathbf{K} = \mathbf{z}_a, \mathbf{V} = \mathbf{z}_a).
\end{equation}
Audio is therefore selected not by raw visual similarity but through 
text, which determines which audio content is relevant at each 
visual position.
The visual bottleneck $\mathbf{z}_v$, the text-aware visual context 
$\mathbf{c}_t^v$, and the text-bridged audio context $\mathbf{c}_a$ are 
then combined through Gated Semantic Modulation (GSM), the 
core block of the adapter (Sec. \ref{sec:gsm}):
\begin{equation}
\mathbf{z}_v' = \text{GSM}(\mathbf{z}_v, \mathbf{c}_a, \mathbf{c}_t^v).
\end{equation}
The audio stream follows the same procedure with 
$\mathbf{T}_a$, producing $\mathbf{c}_t^a$, $\mathbf{c}_v$, and 
$\mathbf{z}_a' = \text{GSM}(\mathbf{z}_a, \mathbf{c}_v, \mathbf{c}_t^a)$. 
The output is projected back to the 
original feature space and added residually:
\begin{equation}
\mathbf{V}^{l+1} = \mathbf{V}^l + \mathbf{z}_v' \mathbf{W}_{\text{up}}^v,
\end{equation}
\begin{equation}
\mathbf{A}^{l+1} = \mathbf{A}^l + \mathbf{z}_a' \mathbf{W}_{\text{up}}^a.
\end{equation}
Cross-modal interaction within each adapter is thus channeled entirely 
through text. Text mediates audio-visual retrieval, contributes directly to each stream's representation, and gates the strength of injection..

\label{sec:method}
\begin{figure}[t]
    \centering
    \scalebox{0.9}{
    \includegraphics[width=\linewidth]{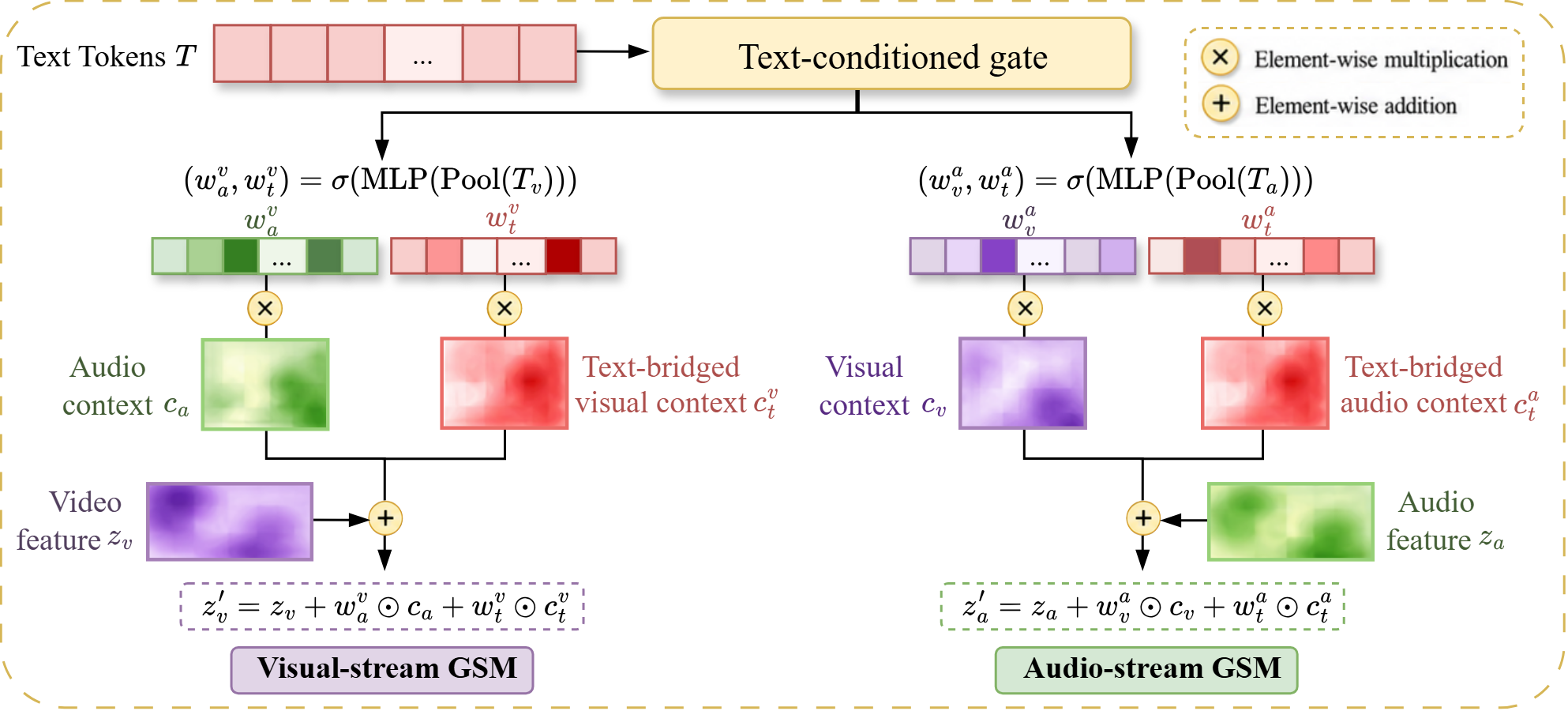}}
    \caption{Gated semantic modulation (GSM). 
The text embedding ($\mathbf{T}_v$ or $\mathbf{T}_a$) drives a gating 
network that decides, per channel, how much text-bridged context 
($\mathbf{c}_t^v$, $\mathbf{c}_t^a$) and cross-modal context 
($\mathbf{c}_a$, $\mathbf{c}_v$) should enter the bottleneck feature 
($\mathbf{z}_v$ or $\mathbf{z}_a$). When text aligns well with a 
channel, the gate amplifies the context; when it does not, 
the gate suppresses it, making text-guided injection adjustable.}
    \label{fig:gsm}
\end{figure}
\subsection{Gated semantic modulation (GSM)}
\label{sec:gsm}

GSM is the core block of TB-AVA that controls how cross-modal context 
is injected into each backbone feature. While the cross-attention in 
Sec.~\ref{sec:tbava} extracts relevant audio and text context for each 
visual position, not all extracted context should influence the backbone equally. 
Text guidance can be unreliable for certain channels, and 
indiscriminately injecting all context drowns useful signal in noise. GSM addresses this by gating each context 
\emph{channel-wise} based on the text embedding, so that text 
selectively determines which channels of the backbone feature receive 
cross-modal updates.

\paragraph{Visual-stream GSM.}
The visual stream's text embedding $\mathbf{T}_v$ is mapped to two 
channel-wise gating vectors via a lightweight gating network:
\begin{equation}
(\mathbf{w}_a^v, \mathbf{w}_t^v) = 
\sigma\!\left(\text{MLP}(\text{Pool}(\mathbf{T}_v))\right),
\end{equation}
where $\mathbf{w}_a^v, \mathbf{w}_t^v \in (0, 1)^{D_b}$ are sigmoid 
gates over the bottleneck channels. The visual bottleneck feature 
$\mathbf{z}_v$ then receives gated injections of the text-bridged 
audio context $\mathbf{c}_a$ and the text-aware visual context 
$\mathbf{c}_t^v$:
\begin{equation}
\mathbf{z}_v' = \mathbf{z}_v 
+ \mathbf{w}_a^v \odot \mathbf{c}_a 
+ \mathbf{w}_t^v \odot \mathbf{c}_t^v,
\end{equation}
where $\odot$ denotes element-wise multiplication broadcast over 
positions. When a context provides reliable evidence on a given channel, the 
corresponding gate approaches one. When text deems the context 
irrelevant, the gate approaches zero, suppressing that channel entirely.

\paragraph{Audio-stream GSM.}
Symmetrically, the audio stream's gates are derived from $\mathbf{T}_a$:
\begin{equation}
(\mathbf{w}_v^a, \mathbf{w}_t^a) = 
\sigma\!\left(\text{MLP}(\text{Pool}(\mathbf{T}_a))\right),
\end{equation}
\begin{equation}
\mathbf{z}_a' = \mathbf{z}_a 
+ \mathbf{w}_v^a \odot \mathbf{c}_v 
+ \mathbf{w}_t^a \odot \mathbf{c}_t^a.
\end{equation}
The two gating networks share the same MLP architecture but use 
modality-specific text embeddings, so that each stream's modulation 
is grounded in the prompt that matches its semantics.

By exposing these gates as explicit, channel-aligned coefficients, 
GSM turns text from an implicit conditioning signal into a controllable 
component of the architecture. The strength of cross-modal injection 
becomes inspectable and modifiable at training and inference time.




\begin{table}[t]
\caption{Performance comparison on the AVE benchmark. 
TB-AVA reaches 85.0\% accuracy with 16.7\% trainable parameters, 
exceeding the strongest prior PEFT method (MoLT, 83.5\%) by 1.5pp.}
  \label{tab:ave_results}
  \centering
  \resizebox{\textwidth}{!}{%
      \begin{tabular}{lccccc}
        \toprule
        \textbf{Method} 
        & \textbf{Visual Encoder} 
        & \textbf{Audio Encoder} 
        & \textbf{Trainable Params (\%)} 
        & \textbf{Total Params (M)} 
        & \textbf{Acc. (\%)} \\
        \midrule
        MBT~\cite{MBT_NeurIPS21} & ViT-B/16 & AST & 100 & 172.0 & 77.8 \\
        \midrule
        AVEL~\cite{tian2018ave} & VGG-19 & VGG-like & 2.7 & 136.0 & 72.7 \\
        CMRAN~\cite{xu2020relationaware} & ResNet-151 & VGG-like & 10.7 & 148.2 & 78.3 \\
        CMBS~\cite{xia2022background} & ResNet-151 & VGG-like & 6.6 & 216.7 & 79.7 \\
        \midrule
        CoPL~\cite{CoPL_MM24} & \multicolumn{2}{c}{ViT-L/16 (shared)} & 1.5 & 332.8 & 79.2 \\
        LAVisH~\cite{LAVisH_CVPR23} & \multicolumn{2}{c}{ViT-L/16 (shared)} & 4.3 & 340.1 & 78.1 \\
        LAVisH~\cite{LAVisH_CVPR23} & \multicolumn{2}{c}{Swin-V2-L (shared)} & 4.1 & 238.8 & 80.8 \\
        AVMoE~\cite{AVMoE_NeurIPS24} & Swin-V2-L & HTS-AT & 34.9 & 404.0 & 80.0 \\
        DG-SCT~\cite{DGSCT_NeurIPS23} & Swin-V2-L & HTS-AT & 43.6 & 461.3 & 81.2 \\
        \rev{Mettle~\cite{DBLP:journals/pami/ZhouLYZGLMHCW26}} & Swin-V2-L & HTS-AT & 23.0 & 338.0 & 83.3 \\
        \rev{MoLT~\cite{rho2025moltmixturelayerwisetokens}}    & Swin-V2-L & HTS-AT & 6.2 & 290.3 & \underline{83.5} \\
        \midrule
        \textbf{TB-AVA without GSM} & SigLIP2-L/16 & BEATs & 16.4 & 486.5 & 83.1 \\
        \textbf{TB-AVA (\textit{Ours})} & SigLIP2-L/16 & BEATs & 16.7 & 488.1 & \textbf{85.0} \\
        \bottomrule
      \end{tabular}%
    }
\end{table}

\begin{figure}[t]
    \centering
    \includegraphics[width=\linewidth]{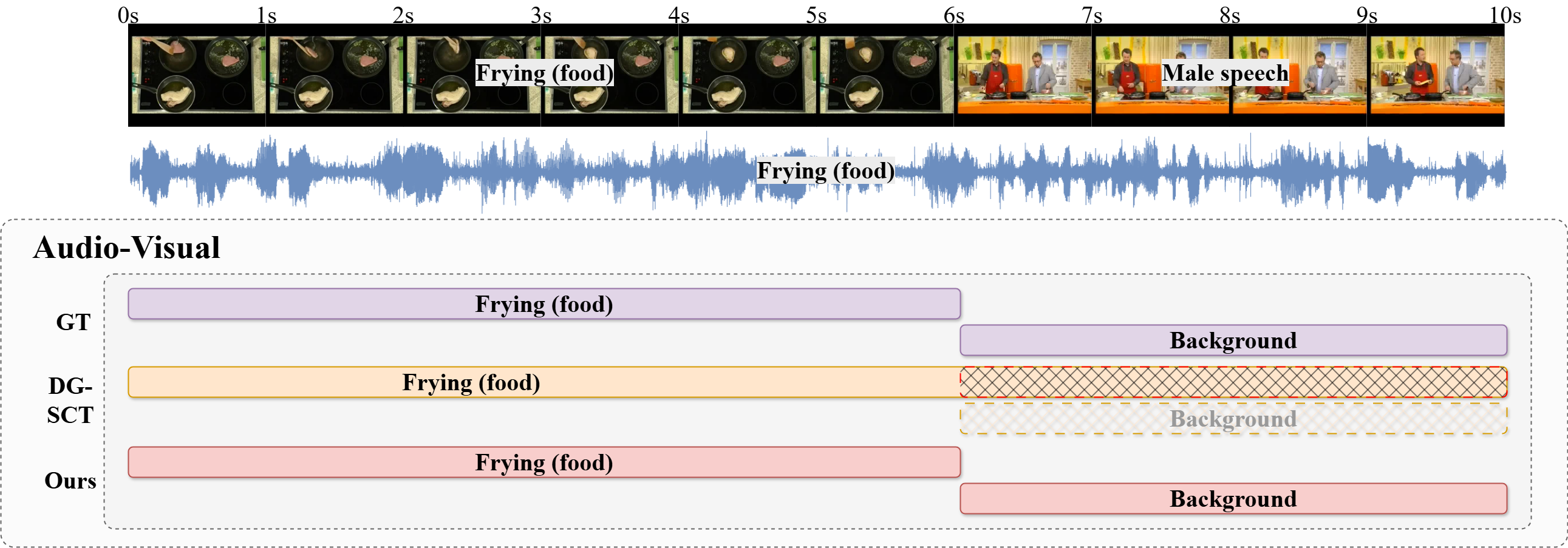}
\caption{Qualitative AVE results. 
TB-AVA correctly assigns \textit{frying (food)} to the first six 
seconds and \textit{background} afterward, while DG-SCT extends 
\textit{frying} across the full clip due to acoustic similarity. 
Text mediation enables prediction changes at semantic boundaries 
that co-occurrence misses.}
    \label{fig:qualitative_ave}
\end{figure}

\begin{table*}[t]
    \caption{Performance comparison on the AVVP benchmark. 
Segment- and event-level F1 on LLP. TB-AVA leads on six of ten 
metrics (V, AV, Type at both granularities), confirming that 
text mediation particularly benefits modality-assignment-sensitive 
predictions under weak video-level supervision. 
MoLT~\cite{rho2025moltmixturelayerwisetokens} is omitted as its 
original paper does not report AVVP results.}
    \label{tab:avvp_results}
    \centering
    \small
    \resizebox{\textwidth}{!}{%
        \begin{tabular}{lcccccccccc}
            \toprule
             & \multicolumn{5}{c}{\textbf{Segment-level}} 
             & \multicolumn{5}{c}{\textbf{Event-level}} \\
            \textbf{Method} 
            & A & V & AV & Type & Event
            & A & V & AV & Type & Event \\
            \midrule
            AVE~\cite{tian2020unified} 
            & 49.9 & 37.3 & 37.0 & 41.4 & 43.6
            & 43.6 & 32.4 & 32.6 & 36.2 & 37.4 \\
            
            HAN~\cite{zhou2023pseudolabel} 
            & 60.1 & 52.9 & 48.9 & 54.0 & 55.4
            & \underline{51.3} & 48.9 & 43.0 & 47.7 & 48.0 \\
            
            MGN~\cite{gao2023presenceabsence} 
            & \underline{60.7} & 55.5 & 50.6 & 55.6 & 57.2
            & 51.0 & 52.4 & 44.4 & 49.3 & 49.2 \\
            
            DG-SCT~\cite{DGSCT_NeurIPS23} 
            & 59.0 & \underline{59.4} & 52.8 & 57.1 & 57.0
            & 49.2 & \underline{56.1} & 46.1 & 50.5 & 49.1 \\

            AVMoE~\cite{AVMoE_NeurIPS24} 
            & 60.5 & 58.9 & \underline{53.2} & \underline{57.5} & \underline{57.8}
            & 50.8 & 54.9 & \underline{46.3} & 50.7 & \underline{49.4} \\
   
            \rev{Mettle~\cite{DBLP:journals/pami/ZhouLYZGLMHCW26}}
            & \textbf{64.3} & 55.9 & 50.9 & 57.0 & \textbf{61.2}
            & \textbf{56.1} & 52.2 & 45.1 & \underline{51.1} & \textbf{53.1} \\
   
            \midrule
            \textbf{TB-AVA (Ours)}
            & 56.4 & \textbf{61.1} & \textbf{55.6} & \textbf{57.7} & 54.9 
            & 48.3 & \textbf{56.7} & \textbf{48.6} & \textbf{51.2} & 47.9 \\
            \bottomrule
        \end{tabular}%
    }
\end{table*}

\begin{figure}[t]
\centering
\begin{minipage}[t]{0.56\textwidth}
  \centering
\captionof{table}{Performance comparison on the AVS benchmark. 
Mean IoU (\%) on AVSBench-object. TB-AVA reaches 81.2 on S4, 
proving competitive for pixel-level audio-visual association.}
  \label{tab:avs_results}
  \vspace{2pt}
  \small                                
  \setlength{\tabcolsep}{5pt}
  \begin{tabular}{lcccc}
    \toprule
    \textbf{Method} & \textbf{Visual} & \textbf{Audio} & \textbf{S4} & \textbf{MS3} \\
    \midrule
    AVS~\cite{zhou2022avseg}    & PVT-v2          & VGGish & 78.7 & 54.0 \\
    \midrule
    LAVisH~\cite{LAVisH_CVPR23} & \multicolumn{2}{c}{Swin-V2-L (shared)} & 80.1 & 49.8 \\
    DG-SCT~\cite{DGSCT_NeurIPS23}    & Swin-V2-L  & HTS-AT & 80.9 & 52.9 \\
    AVMoE~\cite{AVMoE_NeurIPS24}     & Swin-V2-L  & HTS-AT & \underline{81.1} & 53.4 \\
    \rev{Mettle~\cite{DBLP:journals/pami/ZhouLYZGLMHCW26}} & Swin-V2-L & HTS-AT & 80.7 & \underline{55.1} \\
    \rev{MoLT~\cite{rho2025moltmixturelayerwisetokens}}    & Swin-V2-L & HTS-AT & 80.8 & \textbf{57.1} \\
    \midrule
    \textbf{TB-AVA (\textit{Ours})}  & SigLIP2-L/16 & BEATs  & \textbf{81.2} & 53.4 \\
    \bottomrule
  \end{tabular}
\end{minipage}%
\hfill
\begin{minipage}[t]{0.41\textwidth}
  \centering
  \captionof{table}{Backbone-only AVE accuracy varies little across encoder choices.}
  \label{tab:backbone_only}
  \vspace{4pt}
  \footnotesize
  \setlength{\tabcolsep}{4pt}
  \renewcommand{\arraystretch}{0.9}   
  \begin{tabular}{llcc}
    \toprule
    \multicolumn{2}{c}{\textbf{Backbone}} & \textbf{Params} & \textbf{Acc.} \\
    \cmidrule(lr){1-2}
    \textbf{Visual} & \textbf{Audio} & (M) & (\%) \\
    \midrule
    \multicolumn{2}{c}{Swin-V2-L (shared)} & 246 & 74.9 \\
    \multicolumn{2}{c}{CLIP-L (shared)}    & 353 & 76.8 \\
    \multicolumn{2}{c}{SigLIP2 (shared)}   & 319 & 76.5 \\
    \midrule
    Swin-V2-L  & HTS-AT    & 334 & \textbf{77.1} \\
    Video Swin & HTS-AT    & 334 & 75.0 \\
    Swin-V2-L & BEATs & 336 & 72.0 \\
    \midrule
    SigLIP2 & AudioCLIP & 357 & 74.9 \\
    SigLIP2 & HTS-AT    & 407 & 76.6 \\
    SigLIP2 & BEATs     & 409 & \underline{77.0} \\
    \bottomrule
  \end{tabular}
\end{minipage}
\end{figure}
\section{Experiments}
\label{sec:experiments}
\subsection{Setup}

We evaluate TB-AVA on three audio-visual understanding tasks:
Audio-Visual Event Localization (AVE)~\cite{tian2018ave},
Audio-Visual Segmentation (AVS)~\cite{zhou2022avs}, and
Audio-Visual Video Parsing (AVVP)~\cite{tian2020unified}.
For AVE, we use the AVE dataset (4{,}143 videos, 28 categories) and report classification accuracy.
For AVVP, we use the LLP dataset (11{,}849 videos, 25 categories) and report segment- and event-level F1 on audio (A), visual (V), and audio-visual (AV) predictions.
For AVS, we use the AVSBench-object benchmark with both single-source (S4, 4{,}932 videos) and multi-source (MS3, 424 videos) subsets covering 23 semantic classes, evaluated by mean Intersection-over-Union (mIoU).
For all three tasks, the full set of class labels for that benchmark 
is used as the text input, identical across samples and across 
training and inference. Learnable soft prompts are prepended to the 
text tokens but the class names themselves remain fixed.

\subsection{Audio-visual event localization}
\label{sec:exp_ave}

Table~\ref{tab:ave_results} reports classification accuracy on AVE across three method categories: a fully fine-tuned baseline (MBT), task-trained AV architectures, and parameter-efficient adapters over frozen backbones. Among adapter-based methods, the highest reported accuracy is $83.5\%$ 
(MoLT~\cite{rho2025moltmixturelayerwisetokens}, on Swin-V2-L+HTS-AT). 
With $16.7\%$ trainable parameters, TB-AVA reaches $85.0\%$ on 
SigLIP2+BEATs, exceeding MoLT by $1.5$pp. \rev{Section~\ref{sec:backbone} 
isolates the contribution of the text-bridged mechanism from the 
encoder choice.}

Figure~\ref{fig:qualitative_ave} provides a qualitative example. The first six seconds contain a frying-food scene with matching frying audio. At the 6-second mark, the visual cuts to a talk-show segment (male speech) and the audio shifts away from frying, so the audio-visual \emph{frying} event ends at the boundary. DG-SCT predicts \emph{frying (food)} for the entire 10-second window: the audio retains low-level acoustic similarity to frying throughout, and the visual stream still co-occurs in time with that audio, a configuration consistent with a single sustained event under co-occurrence-based alignment. TB-AVA assigns \emph{frying (food)} to the first six seconds and \emph{background} to the remainder, matching the ground truth. The text reference, which lists candidate class names independently of the temporal alignment between modalities, provides an external check on whether the two streams continue to support the same class assignment, and changes the prediction at the boundary where they diverge.

\subsection{Audio-visual video parsing}
\label{sec:exp_avvp}
Table~\ref{tab:avvp_results} reports F1 scores on the LLP benchmark across ten metrics covering segment- and event-level predictions for audio (A), visual (V), audio-visual (AV), Type, and Event categories. TB-AVA obtains the highest F1 on six of the ten metrics: V at both segment ($61.1$) and event ($56.7$) levels, AV at both levels ($55.6$ / $48.6$), and Type at both levels ($57.7$ / $51.2$). Compared with AVMoE, the strongest prior PEFT method on these six metrics, TB-AVA improves V by $2.2$ / $1.8$pp and AV by $2.4$ / $2.3$pp. The remaining four metrics (A and Event at both granularities) are best on Mettle, with TB-AVA below by $7.9$ / $7.8$pp on A and $6.3$ / $5.2$pp on Event.

The metrics where TB-AVA leads (V, AV, Type) all require deciding 
which modality each event belongs to. The metrics where Mettle leads 
(A, Event) only test detection within a single modality. 
\citet{DBLP:journals/pami/ZhouLYZGLMHCW26} link this gap to LLP's 
weak supervision: video-level labels say which events occur, but 
not in which modality. TB-AVA's text input helps with exactly this 
ambiguity; it acts as a third reference against which audio and 
visual evidence can be checked, which is why our gains show up 
mainly on the cross-modal metrics.

\subsection{Audio-visual segmentation}
\label{sec:exp_avs}

Table~\ref{tab:avs_results} reports mIoU on the AVSBench-object benchmark. TB-AVA uses the same adapter configuration as in AVE and AVVP with only the task-specific head varying, and contains no segmentation-specific decoder. On the single-source S4 split, TB-AVA reaches $81.2$ mIoU, above AVMoE ($81.1$), MoLT ($80.8$), and Mettle ($80.7$). The result indicates that text-bridged interaction supplies sufficient signal for pixel-level audio-visual association without a dense-prediction component, and that the parameter-efficient adapter generalizes from segment-level classification to pixel-level prediction without architectural modification.

\rev{On the multi-source MS3 split, TB-AVA reaches $53.4$ mIoU, 
on par with AVMoE and below MoLT ($57.1$). MS3 contains $424$ videos 
with multiple concurrent sounding objects, where the shared global 
class list cannot disambiguate which source corresponds to which 
pixel; S4 is unaffected because each clip contains a single source. 
This is a known limitation of fixed-vocabulary text inputs in 
dense-prediction settings.}
\begin{figure}[t]
\centering

\begin{minipage}[t]{0.46\textwidth}
  \vspace*{0pt}
  \centering
  \captionof{table}{Generalization of TB-AVA across diverse encoder pairs. 
Adding TB-AVA yields consistent gains of +3.7 to +8.1pp across 
four heterogeneous backbone combinations, indicating that the 
improvement stems from the text-bridged alignment mechanism itself.}
  \label{tab:ablation_backbone}
  \setlength{\tabcolsep}{3pt}
  \vspace{2pt}
  \small
  \begin{tabular}{llcc}
    \toprule
    \textbf{Visual} & \textbf{Audio} & \textbf{Baseline} & \textbf{+TB-AVA} \\
    \midrule
    CLIP-L            & CLIP-L            & 76.8 & \textbf{80.7} \\
    Swin-V2-L       & Swin-V2-L       & 74.9 & \textbf{79.4} \\
    Swin-V2-L       & HTS-AT          & 77.1 & \textbf{80.8} \\
    \rev{Swin-V2-L} & \rev{BEATs}     & 72.0 & \textbf{80.1} \\
    
    \bottomrule
  \end{tabular}
\end{minipage}%
\hfill
\begin{minipage}[t]{0.52\textwidth}
  \vspace*{0pt}
  \centering
  \begin{subfigure}[b]{0.48\linewidth}
    \centering
    \includegraphics[width=0.9\linewidth]{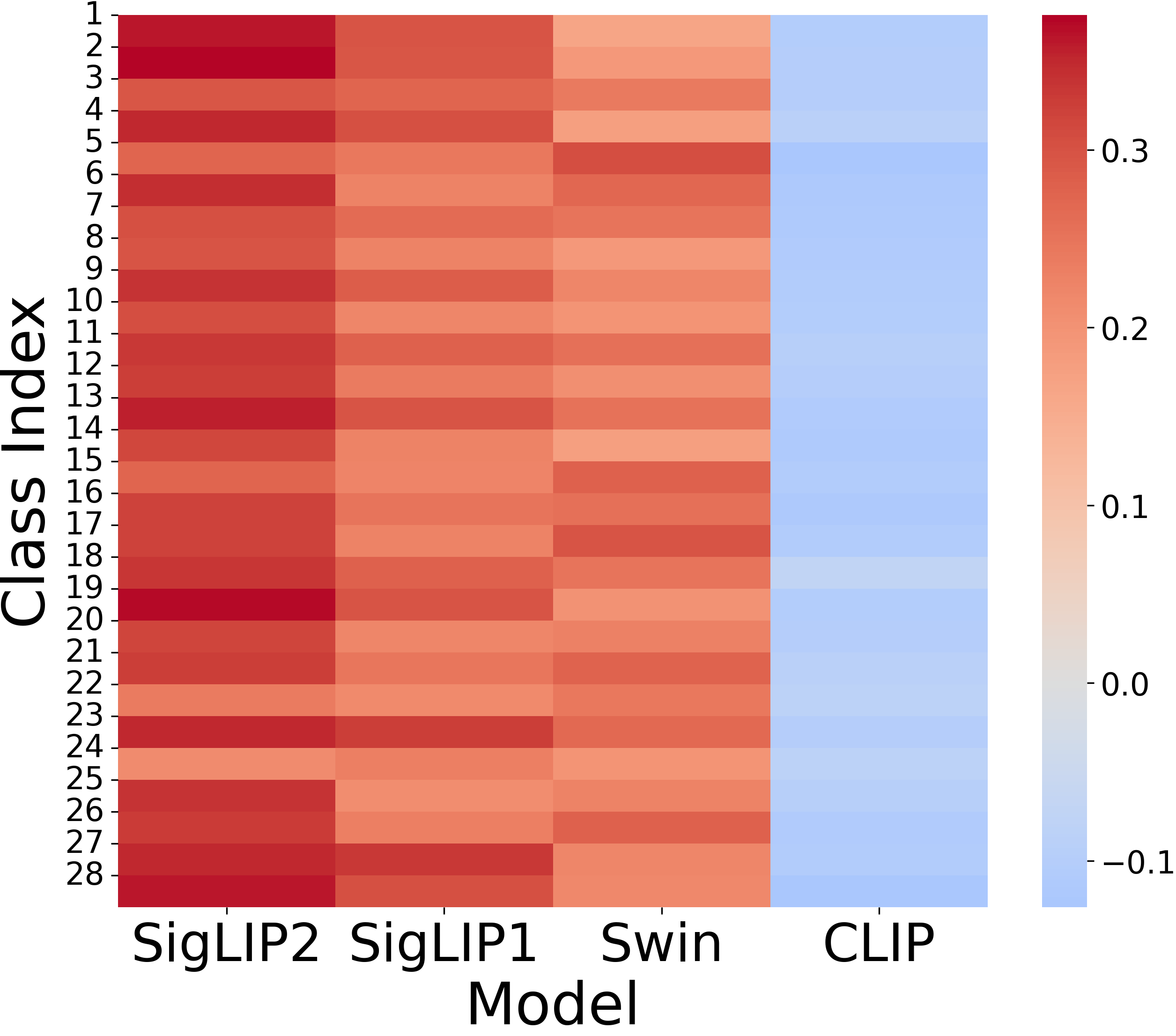}
    \vspace{-1.5mm}
    \caption{Vision-Text alignment}
    \label{fig:vt_heatmap}
  \end{subfigure}%
  \hfill
  \begin{subfigure}[b]{0.48\linewidth}
    \centering
    \includegraphics[width=0.9\linewidth]{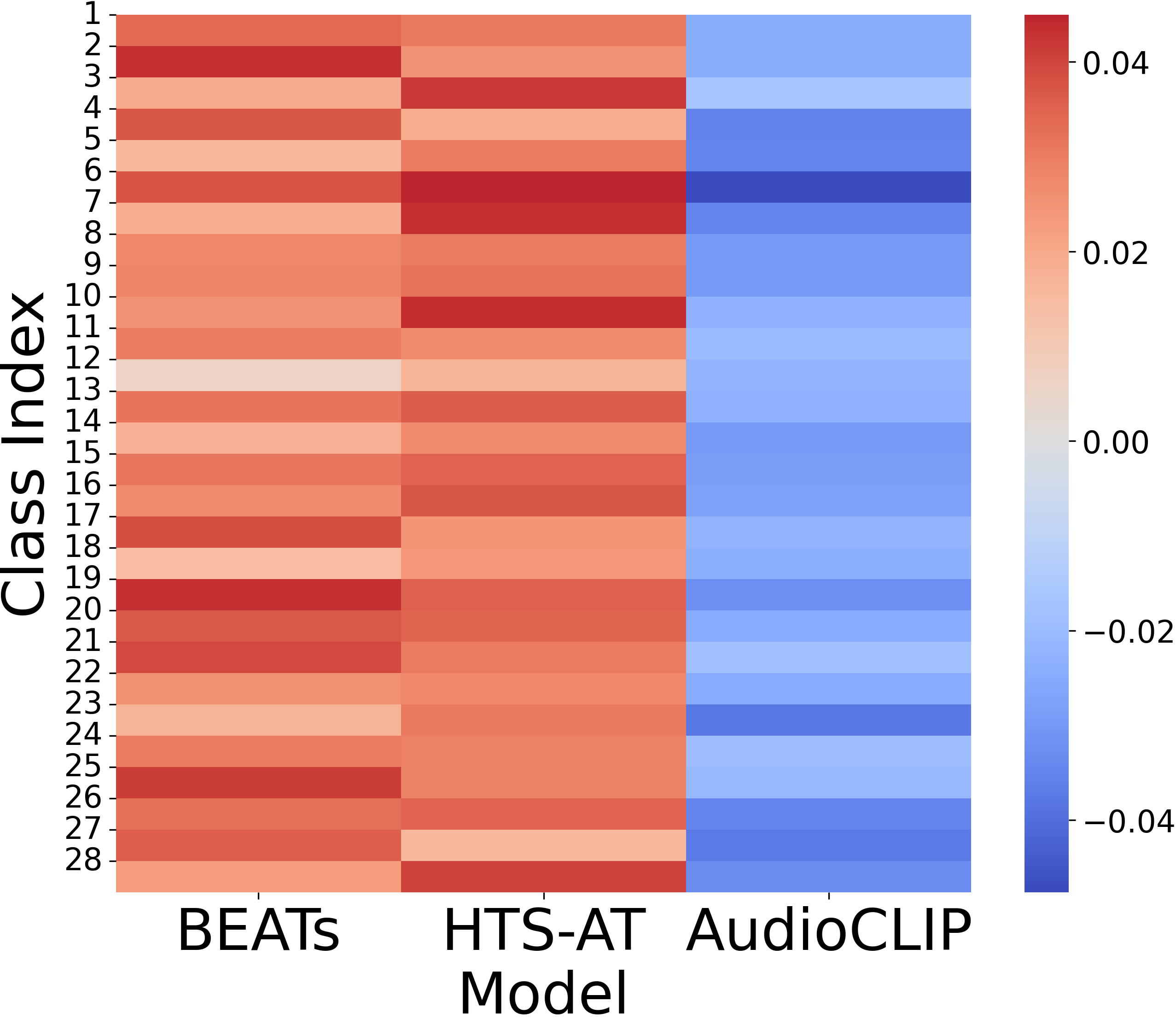}
    \vspace{-1.5mm}
    \caption{Audio-Text alignment}
    \label{fig:at_heatmap}
  \end{subfigure}
  \captionof{figure}{{Class-wise alignment heatmaps in a shared semantic space. 
Cosine similarity between (a) visual and (b) audio features and 
BERT-encoded class embeddings. SigLIP2 and BEATs produce the most 
coherent alignment with text, motivating their use.}}
  \label{fig:alignment_heatmaps}
\end{minipage}

\end{figure}

\section{Analysis}

\subsection{Backbone analysis}
\label{sec:backbone}

Before analyzing the proposed modules, we isolate the contribution of 
the backbones themselves. Table~\ref{tab:backbone_only} reports 
backbone-only accuracy on AVE across eight encoder configurations, 
which fall in a narrow range of $74.9$ to $77.1$\%. Within this range, 
SigLIP2 + BEATs reaches $77.0$\%, on par with Swin-V2-L + HTS-AT 
($77.1$\%): the encoder configuration used by AVMoE, DG-SCT, Mettle, 
and MoLT. Under matched backbone-only conditions, the two 
configurations are therefore comparable, and the improvement to 
$85.0\%$ (Table~\ref{tab:ave_results}) in the full TB-AVA model 
is attributable to the text-bridged adaptation mechanism.

Beyond a single configuration, applying TB-AVA on top of diverse 
encoder combinations yields consistent gains of $+3.7$ to $+8.1$ pp 
(Table~\ref{tab:ablation_backbone}), including architecturally 
heterogeneous pairs such as CLIP+CLIP ($76.8 \rightarrow 80.7$), 
Swin-V2-L+Swin-V2-L ($74.9 \rightarrow 79.4$), Swin-V2-L+HTS-AT 
($77.1 \rightarrow 80.8$), and \rev{Swin-V2-L+BEATs 
($72.0 \rightarrow 80.1$)}. Notably, even the weakest backbone-only 
pair (Swin-V2-L+BEATs at $72.0$\%) closes most of the gap to the 
strongest configurations once TB-AVA is applied, suggesting that 
the text bridge compensates for backbone-level alignment 
mismatches. The absolute ceiling nonetheless differs across 
configurations, and Figure~\ref{fig:alignment_heatmaps} quantifies 
why: cosine similarity between modality features and BERT-encoded 
text embeddings shows that SigLIP2 and BEATs, both pretrained with 
explicit language supervision, produce the most coherent alignment 
patterns. This property motivates their use in our final framework.

\begin{figure}[t]
    \centering
    \scalebox{0.9}{
    \begin{minipage}[b]{0.23\linewidth}
        \begin{subfigure}[b]{\linewidth}
            \centering
            \includegraphics[width=\linewidth]{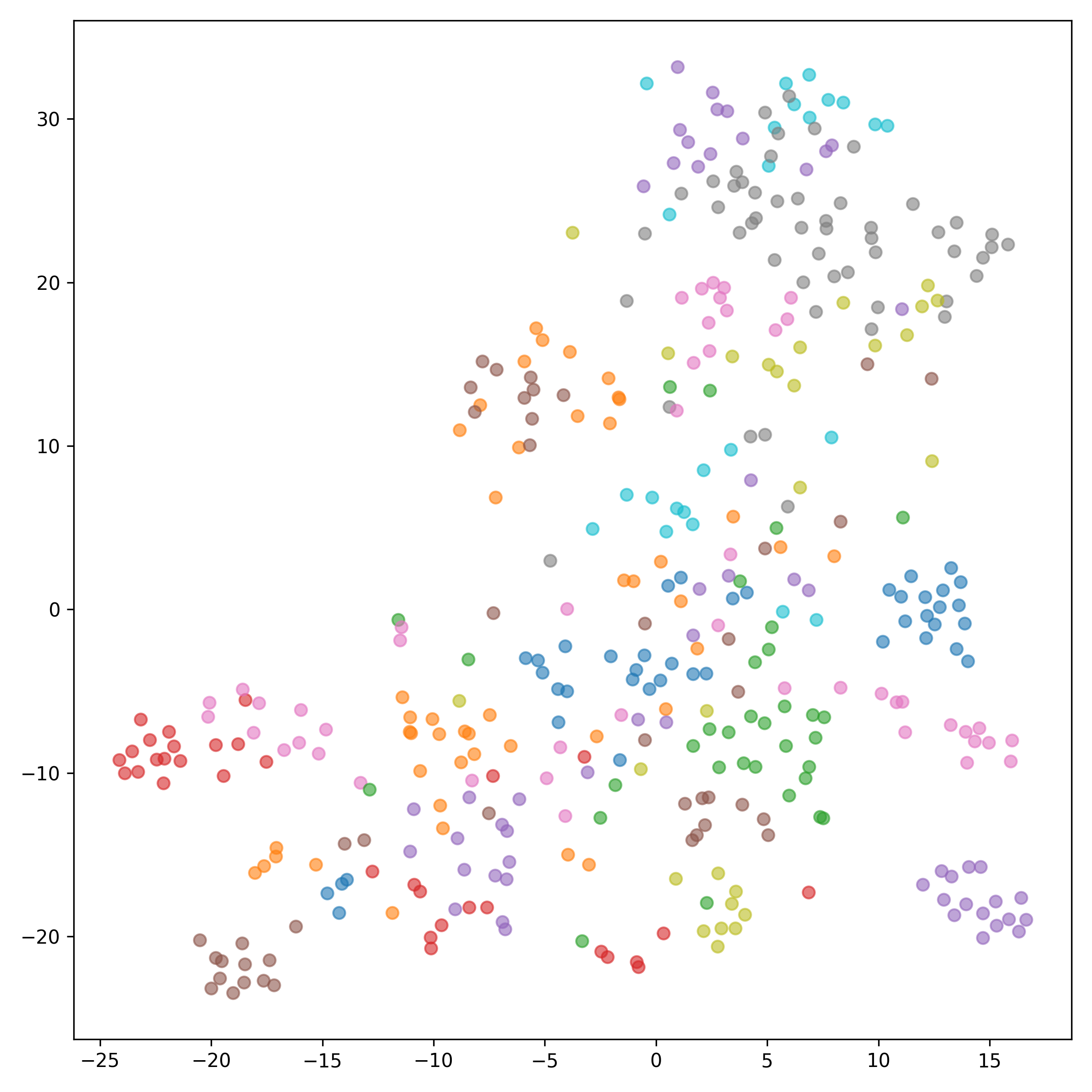}
            \caption{Features without GSM}
            \label{fig:tsne_none}
        \end{subfigure}
        \vfill
        \begin{subfigure}[b]{\linewidth}
            \centering
            \includegraphics[width=\linewidth]{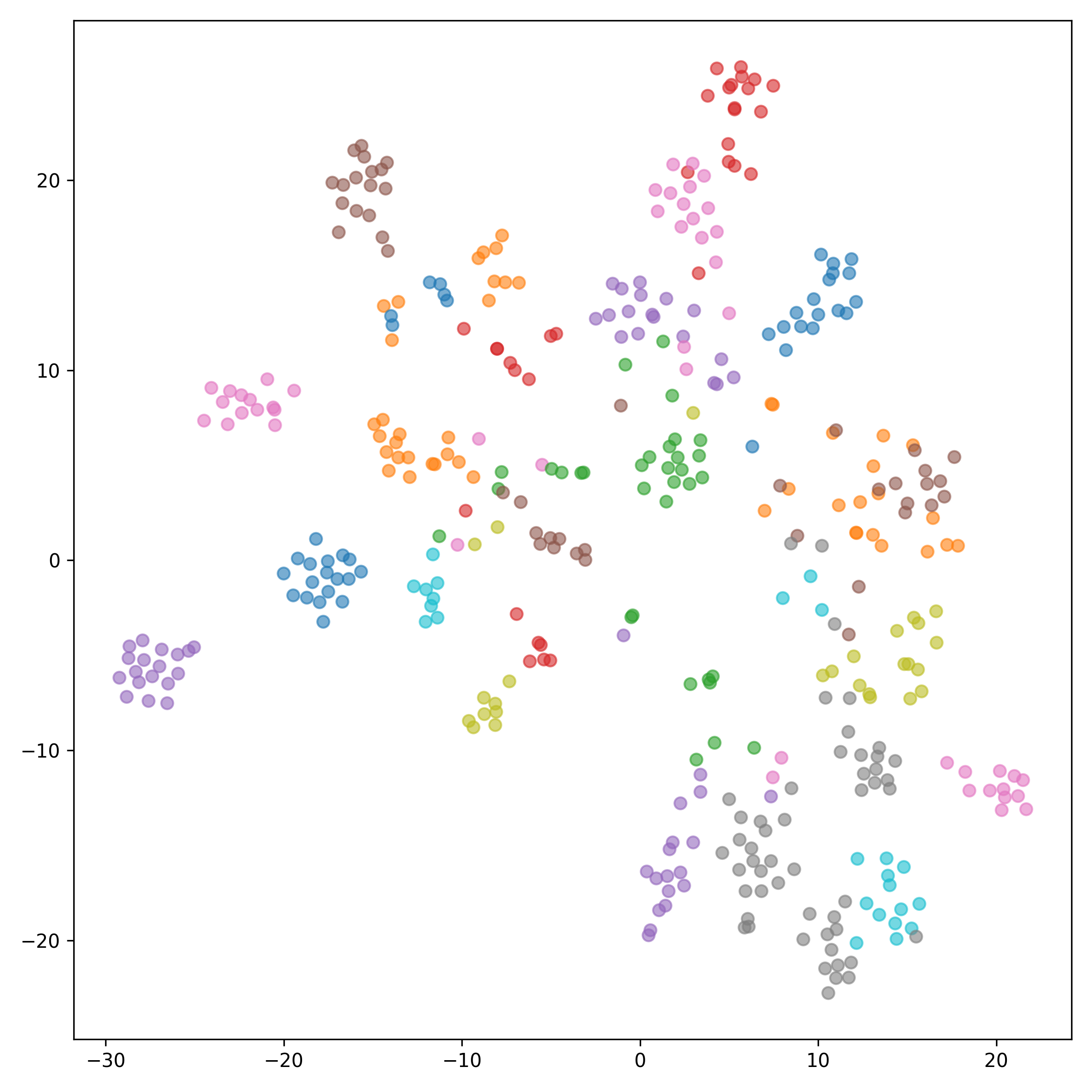}
            \caption{Features with GSM}
            \label{fig:tsne_gated}
        \end{subfigure}
    \end{minipage}
    \hfill
    \begin{minipage}[b]{0.75\linewidth}
        \begin{subfigure}[b]{\linewidth}
            \centering
            \includegraphics[width=\linewidth]{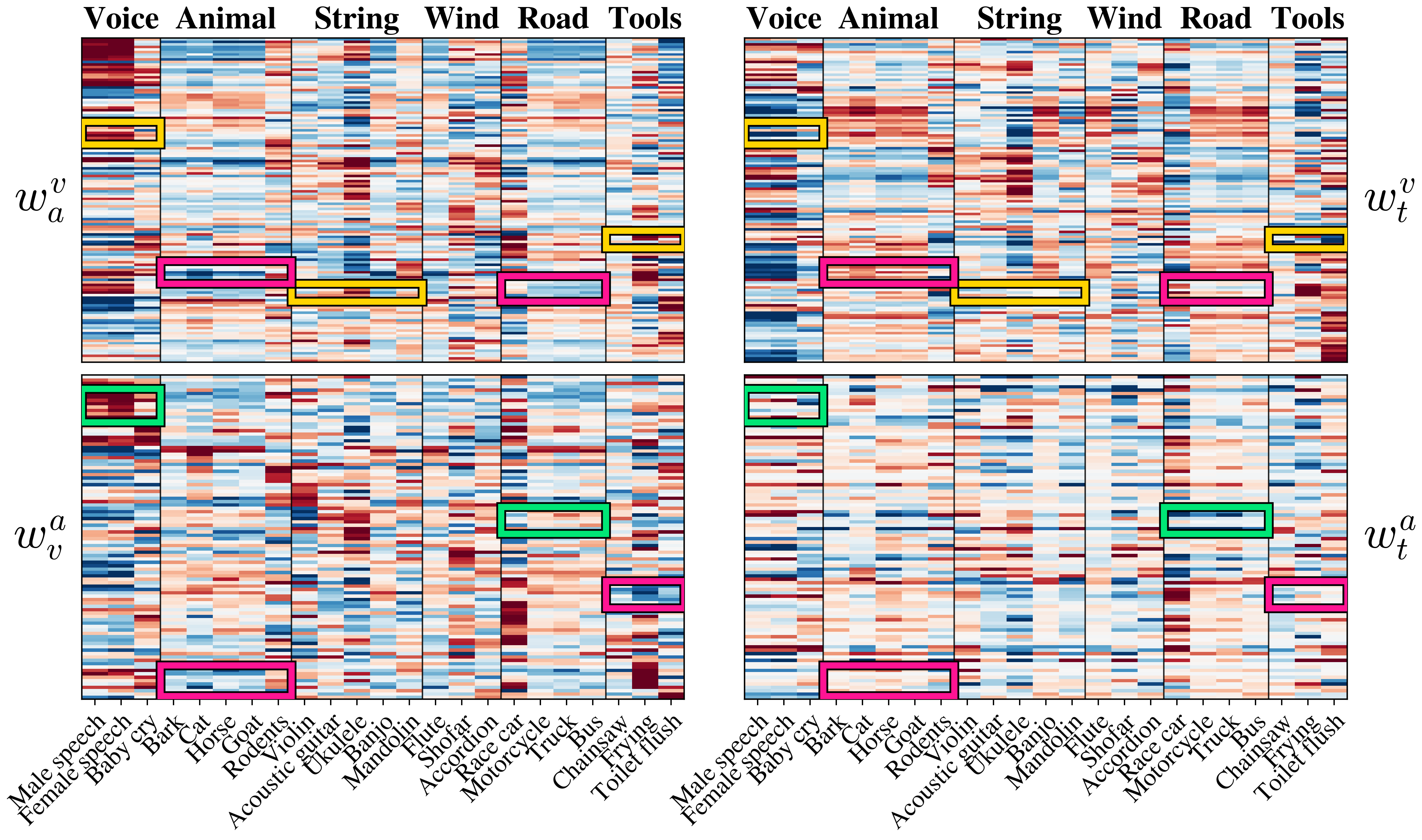}
            \caption{\rev{Channel-wise semantic associations}}
            \label{fig:gsm_heatmap}
        \end{subfigure}
    \end{minipage}}
    \caption{Analysis of GSM on the AVE validation set. 
(a–b) t-SNE projection of joint audio-visual features from the final 
TB-AVA layer, with and without GSM. Adding GSM produces more compact, 
better-separated class-conditional clusters. 
(c) Class-conditional, channel-wise gate values for the visual 
stream (top: $\mathbf{w}_a^v, \mathbf{w}_t^v$) and audio stream 
(bottom: $\mathbf{w}_v^a, \mathbf{w}_t^a$). Rows index bottleneck 
channels; columns index AVE classes grouped by semantic category.}
    \label{fig:gsm_analysis}
\end{figure}

\subsection{Component ablation and behavior of GSM}
\label{sec:ablation}

We conduct ablation studies on the AVE task with frozen SigLIP2 and 
BEATs backbones to isolate the contribution of each module. Starting 
from the backbone-only baseline ($77.0\%$, Table~\ref{tab:backbone_only}), 
replacing GSM with a plain residual addition (the cross-modal adapter 
alone) reaches $83.1\%$, and the full TB-AVA with GSM yields $85.0\%$ 
(Table~\ref{tab:ave_results}). The adapter alone enables text-mediated 
cross-modal interaction, while GSM refines it by selectively 
suppressing channels with weak semantic relevance to the text input.

GSM further refines the cross-modal context through selective rather 
than soft modulation. Compared to TB-AVA without GSM, the full model 
produces more compact and better-separated feature clusters 
(Figure~\ref{fig:gsm_analysis}a--b): class-conditional features no 
longer overlap for visually or acoustically similar categories, and 
within-class variance is reduced even when the underlying audio-visual 
evidence is ambiguous. This suggests that GSM acts as a 
text-conditioned regularizer, sharpening the feature space along 
semantically meaningful directions rather than uniformly modulating 
all channels.

\rev{Per-channel gate analysis (Figure~\ref{fig:gsm_analysis}c) 
further reveals how GSM allocates context across the bottleneck 
channels. Each channel of the bottleneck has a separate gate value 
for each context type (audio/visual cross-modal context and text 
context); a high gate means that channel takes information from 
that context. Plotting these gate values across classes shows 
coherent horizontal bands: groups of channels that consistently 
prefer one context type across an entire class group, rather than 
mixing all contexts uniformly. Three patterns stand out. 
In the visual stream, one band of channels (\textbf{yellow}) prefers 
the audio context $\mathbf{w}_a^v$ on acoustically rich classes 
such as Voice, String, and Tools. In the audio stream, another band 
(\textbf{green}) prefers the visual context $\mathbf{w}_v^a$ on 
visually distinctive classes such as Voice and Road. In both streams, 
a third band (\textbf{pink}) prefers the text context 
($\mathbf{w}_t^v, \mathbf{w}_t^a$) on classes whose modality-specific 
signature is weak, such as Animal. Voice activates in both cross-modal 
directions because speech provides complementary cues in audio 
(timing, prosody) and visual (lip movement, speaker identity). 
This channel-level specialization emerges without supervision, 
supporting GSM's role as a semantic regulator.
}

\section{Conclusion}
\label{sec:conclusion}
We introduced Text-Bridged Audio-Visual Adaptation (TB-AVA), a 
parameter-efficient framework that 1) decouples cross-modal 
alignment from backbone training by routing audio-visual interaction 
through frozen text embeddings, and 2) surpasses fully fine-tuned and adapter-based counterparts 
across AVE, AVVP, and AVS within a parameter-efficient regime. At its core, gated semantic 
modulation (GSM) turns text guidance into an inspectable, 
channel-aligned coefficient, with the two gates spontaneously 
specializing into language-grounded and co-occurrence-grounded roles 
without supervision. Consistent gains of $+3.7$ to $+8.1$pp across 
heterogeneous backbone pairs further indicate that the improvement 
stems from the text-bridged mechanism itself, not from a specific 
encoder choice. We hope these findings encourage further use of text 
as a structural primitive for cross-modal grounding beyond temporal 
co-occurrence.



\medskip
{
\small
\bibliographystyle{plainnat}
\bibliography{example_paper}
}

\end{document}